\documentstyle[fullname]{article}
\author{Ido Dagan \\ Fernando Pereira\\
AT\&T Bell Laboratories \\ 600 Mountain
Ave., Murray Hill, NJ 07974, USA \\{\tt dagan@research.att.com}
\\{\tt pereira@research.att.com}\\
Lillian Lee \\
Division of Applied Sciences, Harvard
University \\  33 Oxford St. Cambridge MA 02138, USA\\
{\tt llee@das.harvard.edu}}
\title{Similarity-Based Estimation of Word Cooccurrence Probabilities
\thanks{To appear in the proceedings of the 32nd Annual Meeting of the
Association for Computational Linguistics, New Mexico State University,
June 1994.}}
\newcommand{\eqpunc}[1]{{\makebox[0pt][l]{\qquad\rm{#1}}}}
\newcommand{\comment}[1]{}
\newcommand{\smrm}[1]{\mbox{\scriptsize #1}}
\newcommand{\todo}[1]{}
\begin{document}
\maketitle
\bibliographystyle{fullname}
\begin{abstract}
In many applications of natural language processing it is
necessary to determine the likelihood of a given word combination. For
example, a speech recognizer may need to determine which of the two
word combinations ``eat a peach'' and ``eat a beach'' is more likely.
Statistical NLP methods determine the likelihood of a word combination
according to its frequency in a training corpus. However, the
nature of language is such that many word combinations are infrequent
and do not occur in a given corpus. In
this work we propose a method for estimating the probability of such
previously unseen word combinations using available information on
``most similar'' words.

We describe a probabilistic word association model based on
distributional word similarity, and apply it to improving probability
estimates for unseen word bigrams in a variant of Katz's back-off
model. The similarity-based method yields a 20\% perplexity improvement
in the prediction of unseen bigrams and statistically
significant reductions in speech-recognition error.
\end{abstract}

\section{Introduction}
\label{introduction}

Data sparseness is an inherent problem in statistical methods for
natural language processing. Such methods use
statistics on the relative frequencies of configurations of elements
in a training corpus to evaluate alternative
analyses or interpretations of new samples of text or speech. The most
likely analysis will be taken to be the one that contains the most
frequent configurations.  The problem of data sparseness arises when
analyses contain configurations that never
occurred in the training corpus. Then it is not possible to
estimate probabilities from observed frequencies, and some other
estimation scheme has to be used.

We focus here on a particular kind of configuration, {\em word
cooccurrence}. Examples of such cooccurrences include relationships
between head words in syntactic constructions (verb-object or
adjective-noun, for example) and word sequences ($n$-grams).  In
commonly used models, the probability estimate for a previously unseen
cooccurrence is a function of the probability estimates for the words
in the cooccurrence. For example, in the bigram models that we study
here, the probability $P(w_2 | w_1)$ of a {\em conditioned word} $w_2$
that has never occurred in training following the {\em conditioning
word} $w_1$ is calculated from the probability of $w_2$, as estimated
by $w_2$'s frequency in the corpus
\cite{Jelinek92,Katz:backoff}.  This method depends on an
independence assumption on the cooccurrence of $w_1$ and $w_2$: the
more frequent $w_2$ is, the higher will be the estimate of $P(w_2 |
w_1)$, regardless of $w_1$.

Class-based and similarity-based models provide an alternative to the
independence assumption. In those models, the relationship between
given words is modeled by analogy with other words that are in some
sense similar to the given ones.

\namecite{Brown+al:class}
suggest a class-based $n$-gram model in which words with similar cooccurrence
distributions are clustered in word classes.  The cooccurrence
probability of a given pair of words then is estimated according to an
averaged cooccurrence probability of the two corresponding classes.
\namecite{Pereira+Tishby+Lee:93} propose a ``soft'' clustering scheme
for certain grammatical cooccurrences in which
membership of a word in a class is probabilistic.  Cooccurrence
probabilities of words are then modeled by averaged cooccurrence
probabilities of word clusters.

\namecite{Dagan+al:contextual} argue that reduction to a relatively small
number of
predetermined word classes or clusters may cause a substantial loss of
information.  Their similarity-based model avoids clustering
altogether.  Instead, each word is modeled by its own specific class,
a set of words which are most similar to it (as in {\em k-nearest
neighbor} approaches in pattern recognition).  Using this scheme, they
predict which unobserved cooccurrences are more likely than
others. Their model, however, is not probabilistic, that is, it does
not provide a probability estimate for unobserved cooccurrences. It
cannot therefore be used in a complete probabilistic framework, such
as $n$-gram language models or probabilistic lexicalized grammars
\cite{Schabes:SLTAG,Lafferty:slink}.

We now give a similarity-based method for estimating the probabilities
of cooccurrences unseen in training.  Similarity-based estimation was
first used for language modeling in the {\em cooccurrence smoothing}
method of \namecite{Essen:92}, derived from work on acoustic model
smoothing by
\namecite{Sugawara+al-85:smoothing}.
We present a different method that takes as starting point the
back-off scheme of \namecite{Katz:backoff}.  We first allocate an
appropriate probability mass for unseen cooccurrences following the
back-off method.  Then we redistribute that mass to unseen
cooccurrences according to an averaged cooccurrence distribution of a
set of most similar conditioning words, using relative entropy as our
similarity measure. This second step replaces the use of the
independence assumption in the original back-off model.

We applied our method to estimate unseen bigram probabilities for {\em
Wall Street Journal} text and compared it to the standard back-off
model.  Testing on a held-out sample, the similarity model achieved a
20\% reduction in perplexity
for unseen bigrams. These constituted
just 10.6\% of the test sample, leading to an overall reduction in
test-set perplexity of 2.4\%. We also experimented with an application
to language modeling for speech recognition, which yielded a
statistically significant reduction in recognition error.

The remainder of the discussion is presented in terms of bigrams, but
it is valid for other types of word cooccurrence as well.

\section{Discounting and Redistribution}
Many low-probability bigrams will be missing from any finite
sample. Yet, the aggregate probability of all these unseen bigrams is
fairly high; any new sample is very likely to contain some.

Because of data sparseness, we cannot reliably use a {\em maximum
likelihood estimator (MLE)} for bigram probabilities.  The MLE for the
probability of a bigram $(w_1,w_2)$ is simply:
\begin{equation}
P_{ML}(w_1,w_2) = \frac{c(w_1,w_2)}{N}\eqpunc{,}
\end{equation}
where $c(w_1,w_2)$ is the frequency of $(w_1,w_2)$ in the training
corpus and $N$ is the total number of bigrams. However,
this estimates the probability of any unseen bigram to be zero, which is
clearly undesirable.

Previous proposals to circumvent the above problem
\cite{Good:53,Jelinek92,Katz:backoff,Church+Gale:GT} take the MLE as
an initial estimate and adjust it so that the total probability of
seen bigrams is less than one, leaving some probability mass for
unseen bigrams. Typically, the adjustment involves either {\em
interpolation}, in which the new estimator is a weighted combination
of the MLE and an estimator that is guaranteed to be nonzero for
unseen bigrams, or {\em discounting}, in which the MLE is decreased
according to a model of the unreliability of small frequency counts,
leaving some probability mass for unseen bigrams.

The back-off model of \namecite{Katz:backoff} provides a clear
separation between frequent events, for which observed frequencies are
reliable probability estimators, and low-frequency events, whose
prediction must involve additional information sources. In addition,
the back-off model does not require complex estimations for
interpolation parameters.

A back-off model requires methods for (a) {\em discounting} the
estimates of previously observed events to leave out some positive
probability mass for unseen events, and (b) {\em
redistributing} among the unseen events the probability mass freed by
discounting. For bigrams the resulting estimator has the
general form
\begin{equation}
\hat{P}(w_2| w_1) = \left\{\!\!
\begin{array}{l@{\hspace{0.7ex}}l}
P_d(w_2 | w_1) & \mbox{if $c(w1,w2) > 0$} \\
\alpha(w_1)P_r(w_2 | w_1) & \mbox{otherwise}
\end{array}\right.\;,\label{genmodel}
\end{equation}
\noindent where $P_d$ represents the discounted estimate for seen
bigrams, $P_r$ the model for probability redistribution among the
unseen bigrams, and $\alpha(w)$ is a normalization factor. Since the
overall mass left for unseen bigrams starting with $w_1$ is given by
\[
\tilde{\beta}(w_1) = 1 - \sum_{w_2:c(w_1,w_2) > 0}
P_d(w_2 | w_1) \eqpunc{,}
\]
the normalization factor required to ensure
$\sum_{w_2}\hat{P}(w_2| w_1) = 1$ is
\begin{eqnarray*}
\alpha(w_1) & = &\frac{\tilde{\beta}(w_1)}{\sum_{w_2:c(w_1,w_2)=
0} P_r(w_2 | w_1)} \\
& = & \frac{\tilde{\beta}(w_1)}{1-\sum_{w_2:c(w_1,w_2)>0} P_r(w_2 |
w_1)}\eqpunc{.}
\end{eqnarray*}
The second formulation of the normalization is computationally
preferable because the total number of possible bigram types far exceeds
the number of observed types. Equation (\ref{genmodel}) modifies
slightly Katz's presentation to include the placeholder $P_r$ for
alternative models of the distribution of unseen bigrams.

Katz uses the Good-Turing formula to replace the actual
frequency $c(w_1,w_2)$ of a bigram (or
an event, in general) with a discounted frequency, $c^{*}(w_1,w_2)$,
defined by
\begin{equation}
\label{discount}
c^{*}(w_1,w_2) = (c(w_1,w_2) + 1) \frac{n_{c(w_1,w_2) + 1}}{n_{c(w_1,w_2)}}\;,
\end{equation}
where $n_c$ is the number of different bigrams in the corpus that have
frequency $c$.
He then uses the discounted frequency in the conditional probability
calculation for a bigram:
\begin{equation}
\label{gt}
P_{d}(w_2|w_1) = \frac{c^*(w_1,w_2)}{c(w_1)}\eqpunc{.}
\end{equation}

In the original Good-Turing method \cite{Good:53} the free
probability mass is redistributed uniformly among all unseen events.
Instead, Katz's back-off scheme redistributes the free probability
mass non-uniformly in proportion to the frequency of $w_2$, by setting
\begin{equation}
P_r(w_2|w_1) = P(w_2)\eqpunc{.}\label{ubo}
\end{equation}
Katz thus assumes that for a given conditioning word $w_1$ the
probability of an unseen following word $w_2$ is proportional to its
unconditional probability. However, the overall form of the model
(\ref{genmodel}) does not depend on this assumption, and we will next
investigate an estimate for $P_r(w_2| w_1)$ derived by averaging
estimates for the conditional probabilities that $w_2$ follows words
that are distributionally similar to $w_1$.

\section{The Similarity Model}

Our scheme is based on the assumption that words that are ``similar'' to
$w_1$ can provide good predictions for the
distribution of $w_1$ in unseen bigrams.
Let ${\cal S}(w_1)$ denote a set of words which are most similar to
$w_1$, as determined by some similarity metric.
We define $P_{\smrm{SIM}}(w_2|w_1)$, the similarity-based model for the
conditional distribution of $w_1$,
as a weighted average of the conditional distributions of the words in
${\cal S}(w_1)$:
\begin{equation}
\label{Psim}
\begin{array}{l}
\lefteqn{P_{\smrm{SIM}}(w_2|w_1) =} \\
\qquad\sum_{w'_{1} \in {\cal S}(w_1)} P(w_2|w'_{1}) \frac{
           W(w'_{1},w_1)}{
        \sum_{w'_{1} \in {\cal S}(w_1)}
           W(w'_{1},w_1)}
\end{array}\;,
\end{equation}
where $W(w'_{1},w_1)$ is the (unnormalized) weight given to $w'_{1}$,
determined by its degree of similarity to $w_1$.
According to this scheme, $w_2$ is more likely to follow $w_1$ if it
tends to follow words that are most similar to $w_1$.
To complete the scheme, it is necessary to define the similarity
metric and, accordingly, ${\cal S}(w_1)$ and $W(w'_{1},w_1)$.

Following \namecite{Pereira+Tishby+Lee:93}, we
measure word similarity by the
{\em  relative entropy}, or {\em Kullback-Leibler (KL) distance},
between the corresponding conditional distributions
\begin{equation}
\label{KL}
D(w_1 \parallel w'_{1}) = \sum_{w_2} P(w_2|w_1)
              \log \frac{P(w_2|w_1)}{P(w_2|w'_{1})}\;.
\end{equation}
The KL distance is 0 when $w_1 = w'_{1}$, and it increases as the
two distribution are less similar.

To compute (\ref{Psim}) and (\ref{KL}) we must have nonzero estimates
of $P(w_2|w_1)$ whenever necessary for (\ref{KL}) to be defined. We
use the estimates given by the standard back-off model, which satisfy
that requirement. Thus our application of the similarity model
averages together standard back-off estimates for a set of similar
conditioning words.

We define
${\cal S}(w_1)$  as the set of
at most $k$ nearest words to $w_1$ (excluding $w_1$ itself), that
also satisfy $D(w_1 \parallel w'_{1})<t$.
$k$ and $t$ are parameters that control the contents of ${\cal
S}(w_1)$ and are tuned experimentally, as we will see below.

$W(w'_{1},w_1)$ is defined as
\[
W(w'_{1},w_1) = \exp - \beta D(w_1 \parallel w'_{1})\eqpunc{.}
\]
The weight is larger for words that are
more similar (closer) to $w_1$. The parameter
$\beta$ controls the relative contribution of
words in different distances from $w_1$: as the value  of $\beta$
increases, the nearest words to $w_1$ get relatively more weight.
As $\beta$ decreases, remote words get a larger effect. Like $k$ and
$t$, $\beta$ is tuned experimentally.

Having a definition for $P_{\smrm{SIM}}(w_2|w_1)$, we could use it
directly as $P_r(w_2|w_1)$ in the back-off scheme (\ref{genmodel}). We
found that it is better to smooth $P_{\smrm{SIM}}(w_2|w_1)$ by
interpolating it with the unigram probability $P(w_2)$ (recall that
Katz used $P(w_2)$ as $P_r(w_2|w_1)$). Using linear interpolation we
get
\begin{equation}
\label{ourPr}
P_r(w_2|w_1) = \gamma P(w_2) + (1 - \gamma) P_{\smrm{SIM}}(w_2|w_1)\;,
\end{equation}
where $\gamma$ is an experimentally-determined interpolation parameter.
This smoothing appears to compensate for
inaccuracies in $P_{\smrm{SIM}}(w_2|w_1)$, mainly for infrequent
conditioning words.
However, as the evaluation below shows, good values for $\gamma$ are
small, that is, the similarity-based model plays a stronger role than
the independence assumption.

To summarize, we construct a similarity-based model for $P(w_2|w_1)$
and then interpolate it with $P(w_2)$.
The interpolated model (\ref{ourPr}) is used in the back-off scheme
as $P_r(w_2|w_1)$, to obtain better estimates for unseen bigrams.
Four parameters, to be tuned experimentally, are relevant for this
process: $k$ and $t$, which determine the set of similar words to be
considered, $\beta$, which determines the relative effect of these
words, and $\gamma$, which
determines the overall importance of the similarity-based model.

\section{Evaluation}
\label{evaluation}
We evaluated our method by comparing its perplexity\footnote{The
perplexity of a conditional bigram probability model $\hat{P}$ with
respect to the true bigram distribution is an
information-theoretic measure of model quality \cite{Jelinek92} that
can be empirically estimated by $\exp -\frac{1}{N} \sum_i \log
\hat{P}(w_i|w_{i-1})$ for a test set of length $N$.
Intuitively, the lower the perplexity of a model the more
likely the model is to assign high probability to bigrams that
actually occur. In our task, lower perplexity will indicate better
prediction of unseen bigrams.}  and effect on speech-recognition
accuracy with the baseline bigram back-off model developed by MIT
Lincoln Laboratories for the {\em Wall Street Journal} (WSJ) text and
dictation corpora provided by ARPA's HLT program
\cite{Paul:91}.%
\footnote{The ARPA WSJ development corpora come in two
versions, one with verbalized punctuation and the other without. We
used the latter in all our experiments.}  The baseline back-off model
follows closely the Katz design, except that for
compactness all frequency one bigrams are ignored. The counts used in
this model and in ours were obtained from
40.5 million words of WSJ text from the years 1987-89.

For perplexity evaluation, we tuned the similarity model parameters
by minimizing perplexity on an
additional sample of 57.5 thousand
words of WSJ text, drawn from the ARPA HLT development test set.
The best parameter values found were
$k=60$, $t=2.5$, $\beta=4$ and
$\gamma=0.15$. For these values, the improvement in perplexity for
unseen bigrams in a held-out 18 thousand word sample, in which 10.6\%
of the bigrams are unseen,
is just over 20\%. This improvement on unseen bigrams
corresponds to an overall test set perplexity improvement of 2.4\%
(from 237.4 to 231.7).
\begin{table*}
\begin{center}
\begin{tabular}{rrrrrr}
$k$ & $t$ & $\beta$ & $\gamma$ & training reduction (\%) & test reduction
(\%)\\
\hline
60 & 2.5 & 4 & 0.15 & 18.4 & 20.51 \\
50 & 2.5 & 4 & 0.15 & 18.38 & 20.45 \\
40 & 2.5 & 4 & 0.2 & 18.34 & 20.03 \\
30 & 2.5 & 4 & 0.25 & 18.33 & 19.76 \\
70 & 2.5 & 4 & 0.1 & 18.3 & 20.53 \\
80 & 2.5 & 4.5 & 0.1 & 18.25 & 20.55 \\
100 & 2.5 & 4.5 & 0.1 & 18.23 & 20.54 \\
90 & 2.5 & 4.5 & 0.1 & 18.23 & 20.59 \\
20 & 1.5 & 4 & 0.3 & 18.04 & 18.7 \\
10 & 1.5 & 3.5 & 0.3 & 16.64 & 16.94
\end{tabular}
\end{center}
\caption{Perplexity Reduction on Unseen Bigrams for Different Model Parameters}
\label{perp-results}
\end{table*}
Table \ref{perp-results} shows reductions in training and test
perplexity, sorted by training reduction, for different choices in the
number $k$ of closest neighbors used. The values of $\beta$, $\gamma$
and $t$ are the best ones found for each $k$.%
\footnote{Values of
$\beta$ and $t$ refer to base 10 logarithms and exponentials in all
calculations.}

{}From equation (\ref{Psim}), it is clear that the computational cost of
applying the similarity model to an unseen bigram is
$O(k)$. Therefore, lower values for $k$ (and
also for $t$) are computationally preferable.
\todo{How many times is the $t$ limit actually invoked?}
{}From the table, we can see that reducing $k$ to 30 incurs
a penalty of less than 1\% in the perplexity improvement, so
relatively low values of $k$ appear to be sufficient to achieve most
of the benefit of the similarity model. As the table also shows, the
best value of $\gamma$ increases as $k$ decreases, that is, for lower
$k$ a greater weight is given to the conditioned word's frequency.
This suggests that the predictive power of neighbors beyond the
closest 30 or so can be modeled fairly well by the overall frequency
of the conditioned word.

The bigram similarity model was also tested as a language model in
speech recognition. The test data for this experiment were pruned word
lattices for 403 WSJ closed-vocabulary test sentences.  Arc scores in those
lattices are sums of an acoustic score (negative log likelihood) and a
language-model score, in this case the negative log probability
provided by the baseline bigram model.

{}From the given lattices, we constructed new lattices in which the arc
scores were modified to use the similarity model instead of the
baseline model.  We compared the best sentence hypothesis in each
original lattice and in the modified one, and counted the word
disagreements in which one of the hypotheses is correct.  There were a
total of 96 such disagreements. The similarity model was correct in 64
cases, and the back-off model in 32. This advantage for the similarity
model is statistically significant at the 0.01 level. The overall
reduction in error rate is small, from 21.4\% to 20.9\%, because the
number of disagreements is small compared with the overall number of
errors in our current recognition setup.

Table \ref{rec-eg} shows some examples of speech recognition
disagreements between the two models. The hypotheses are labeled `B'
for back-off and `S' for similarity, and the bold-face words are
errors. The similarity model seems to be able to model better
regularities such as semantic parallelism in lists and avoiding a past
tense form after ``to.'' On the other hand, the similarity model
makes several mistakes in which a function word is inserted in a place
where punctuation would be found in written text.

\begin{table*}
\begin{center}
\begin{tabular}{l|l}
B & commitments \ldots from leaders {\bf felt the} three point six
billion dollars \\
\hline S & commitments \ldots from leaders fell to three point six
billion dollars \\
\hline \hline B & followed by France the US {\bf
agreed in}  Italy  \\
\hline S & followed by France the US Greece
\ldots Italy \\
\hline\hline B & he whispers to {\bf made a} \\
\hline
S & he whispers to an aide \\
\hline\hline B & the necessity for change
{\bf exist} \\
 \hline S & the necessity for change exists \\
\hline\hline\hline B & without \ldots additional reserves Centrust  would
have reported \\
\hline S & without \ldots additional reserves {\bf
of} Centrust would have reported \\
\hline \hline  B & in the darkness
past the church \\
\hline S & in the darkness {\bf passed} the church
\end{tabular}
\end{center}
\caption{Speech Recognition Disagreements between Models}
\label{rec-eg}
\end{table*}

\section{Related Work}
The {\em cooccurrence smoothing} technique
\cite{Essen:92}, based on earlier stochastic speech modeling work by
\namecite{Sugawara+al-85:smoothing}, is the main previous attempt to
use similarity to estimate the probability of unseen events in
language modeling. In addition to its original use in language
modeling for speech recognition,
\namecite{Grishman+Sterling-93:selectional} applied the cooccurrence
smoothing technique to estimate the likelihood of selectional
patterns.  We will outline here the main parallels and differences
between our method and cooccurrence smoothing. A more
detailed analysis would require an empirical comparison of the two
methods on the same corpus and task.

In cooccurrence smoothing, as in our method, a baseline model is
combined with a similarity-based model that refines some of its
probability estimates. The similarity model in cooccurrence smoothing
is based on the intuition that the similarity between two words $w$
and $w'$ can be measured by the {\em confusion} probability
$P_C(w'|w)$ that $w'$ can be substituted for $w$ in an arbitrary
context in the training corpus. Given a baseline probability model
$P$, which is taken to be the MLE,
the confusion
probability $P_C(w'_{1}|w_1)$ between conditioning words $w'_{1}$ and
$w_1$ is defined as
\begin{equation}
\begin{array}{l}
\lefteqn{P_C(w'_{1}|w_1) =} \\
\qquad \frac{1}{P(w_1)}\sum_{w_2} P(w_1|w_2)
P(w'_{1}|w_2) P(w_2)
\end{array}\;,
\label{cooc-sim}
\end{equation}
the probability that $w_1$ is followed by the same context words as
$w'_{1}$. Then the bigram estimate derived by cooccurrence
smoothing is given by
\[
P_S(w_2|w_1) = \sum_{w'_{1}}P(w_2|w'_{1})
P_C(w'_{1}|w_1)\eqpunc{.}
\]
Notice that this formula has the same form as our similarity model
(\ref{Psim}), except that it uses confusion probabilities where we use
normalized weights.%
\footnote{This presentation corresponds to model 2-B in
\namecite{Essen:92}. Their presentation follows the equivalent model
1-A, which averages over similar
conditioned words, with the similarity defined with the preceding word
as context. In fact, these equivalent models
are symmetric in their treatment of conditioning and conditioned word,
as they can both be rewritten as
\[
P_S(w_2|w_1) = \sum_{w'_{1},w'_{2}} P(w_2|w'_{1})
P(w'_{1}|w'_{2}) P(w'_{2}|w_1) \quad.
\]
They also consider other definitions of confusion
probability and smoothed probability estimate, but the one above yielded
the best experimental results.} In addition, we restrict the summation
to sufficiently similar words, whereas the cooccurrence smoothing
method sums over all words in the lexicon.

The similarity measure (\ref{cooc-sim}) is symmetric in the sense that
$P_C(w'|w)$ and $P_C(w|w')$ are identical up to frequency
normalization, that is $\frac{P_C(w'|w)}{P_C(w|w')} =
\frac{P(w)}{P(w')}$. In contrast, $D(w\parallel w')$ (\ref{KL}) is
asymmetric in that it weighs each context in proportion to its
probability of occurrence with $w$, but not with $w'$.  In this way,
if $w$ and $w'$ have comparable frequencies but $w'$ has a sharper
context distribution than $w$, then $D(w'\parallel w)$ is greater than
$D(w\parallel w')$. Therefore, in our similarity model $w'$ will play a
stronger role in estimating $w$ than vice versa. These properties
motivated our choice of relative entropy for similarity measure, because
of the intuition that words with sharper distributions are more
informative about other words than words with flat distributions.

Finally, while we have used our similarity model only for missing
bigrams in a back-off scheme, \namecite{Essen:92} used linear
interpolation for all bigrams to combine the cooccurrence smoothing
model with MLE models of bigrams and unigrams. Notice, however, that
the choice of back-off or interpolation is independent from the
similarity model used.

\section{Further Research}

Our model provides a basic scheme for probabilistic similarity-based
estimation that can be developed in several directions.  First,
variations of (\ref{Psim}) may be tried, such as different similarity
metrics and different weighting schemes. Also, some simplification of
the current model parameters may be possible, especially with respect
to the parameters $t$ and $k$ used to select the nearest neighbors of
a word.  A more substantial variation would be to base the model on
similarity between conditioned words rather than on similarity between
conditioning words.

Other evidence may be combined with the similarity-based estimate.
For instance, it may be advantageous to weigh those estimates by some
measure of the reliability of the similarity metric and of the
neighbor distributions.  A second possibility is to take into account
negative evidence: if $w_1$ is frequent, but $w_2$ never followed it,
there may be enough statistical evidence to put an upper bound on the
estimate of $P(w_2|w_1)$. This may require an adjustment of the
similarity based estimate, possibly along the lines of
\cite{Rosenfeld+Huang}.  Third, the similarity-based estimate can be
used to smooth the maximum likelihood estimate for small nonzero
frequencies.  If the similarity-based estimate is relatively high, a
bigram would receive a higher estimate than predicted by the uniform
discounting method.

Finally, the similarity-based model may be applied to configurations
other than bigrams.  For trigrams, it is necessary to measure
similarity between different conditioning bigrams. This can be done
directly, by measuring the distance between distributions of the form
$P(w_3|w_1,w_2)$, corresponding to different bigrams
$(w_1,w_2)$. Alternatively, and more practically, it would be possible
to define a similarity measure between bigrams as a function of
similarities between corresponding words in them.  Other types of
conditional cooccurrence probabilities have been used in probabilistic
parsing \cite{Black+al-93:history}.  If the configuration in question
includes only two words, such as $P(object|verb)$, then it is possible
to use the model we have used for bigrams. If the configuration
includes more elements, it is necessary to adjust the method, along
the lines discussed above for trigrams.

\section{Conclusions}
\label{conclusions}

Similarity-based models suggest an appealing approach for dealing with
data sparseness. Based on corpus statistics, they provide analogies
between words that often agree with our linguistic and domain
intuitions.  In this paper we presented a new model that implements
the similarity-based approach to provide estimates for the conditional
probabilities of unseen word cooccurrences.

Our method combines similarity-based estimates with Katz's back-off
scheme, which is widely used for language modeling in speech
recognition. Although the scheme was originally proposed as a
preferred way of implementing the independence assumption, we suggest
that it is also appropriate for implementing similarity-based models,
as well as class-based models.  It enables us to rely on direct
maximum likelihood estimates when reliable statistics are available,
and only otherwise resort to the estimates of an ``indirect'' model.

The improvement we achieved for a bigram model is statistically
significant, though modest in its overall effect because of the small
proportion of unseen events.  While we have used bigrams as an
easily-accessible platform to develop and test the model, more
substantial improvements might be obtainable for more informative
configurations.  An obvious case is that of trigrams, for which the
sparse data problem is much more severe.%
\footnote{For WSJ trigrams, only 58.6\%
of test set trigrams occur in 40M of words of training (Doug Paul,
personal communication).} Our longer-term goal, however, is to apply
similarity techniques to linguistically motivated word cooccurrence
configurations, as suggested by lexicalized approaches to parsing
\cite{Schabes:SLTAG,Lafferty:slink}. In configurations like
verb-object and adjective-noun, there is some evidence
\cite{Pereira+Tishby+Lee:93} that sharper word cooccurrence
distributions are obtainable, leading to improved
predictions by similarity techniques.
\section*{Acknowledgments}
We thank Slava Katz for discussions on the topic of this paper, Doug
McIlroy for detailed comments, Doug Paul for help with his baseline
back-off model, and Andre Ljolje and Michael Riley for providing the
word lattices for our experiments.

\end{document}